\documentclass{article} 

\usepackage[colorlinks=true,citecolor=blue]{hyperref} 
\usepackage{styles/jfrExamplee}
\usepackage{graphicx}
\usepackage{setspace}
\usepackage{epstopdf}
\usepackage{enumitem} 		
\usepackage[tight]{units} 	
\usepackage{amsmath}		
\usepackage{amssymb}		
\usepackage{algorithm2e}	
\usepackage{gensymb}		
\usepackage{booktabs}  		
\usepackage{multirow}		
\usepackage{caption}		
\usepackage{calrsfs}		
\DeclareMathAlphabet{\pazocal}{OMS}{zplm}{m}{n}
\usepackage[toc,page]{appendix}		
\usepackage[figuresleft]{rotating}		
\usepackage{mathtools}		
\usepackage{tablefootnote}	

\usepackage[backend=biber, style=apa]{biblatex}
\usepackage[american]{babel}						
\usepackage{csquotes}								
\DeclareLanguageMapping{american}{american-apa}		
\usepackage{color} 
\usepackage{soul} 
\sethlcolor{green}
\definecolor{OliveGreen}{rgb}{0,0.6,0}

\DeclareMathAlphabet{\pazocal}{OMS}{zplm}{m}{n}
\let\oldnl\nl
\newcommand{\nonl}{\renewcommand{\nl}{\let\nl\oldnl}}

\makeatletter
\newcommand{\nosemic}{\renewcommand{\@endalgocfline}{\relax}}
\newcommand{\dosemic}{\renewcommand{\@endalgocfline}{\algocf@endline}}
\makeatother

\setlist[itemize]{itemsep=0pt,parsep=3pt,partopsep=0pt, topsep=3pt, leftmargin=25pt,rightmargin=25pt}
\setlist[enumerate]{itemsep=0pt,parsep=3pt,partopsep=0pt,topsep=0pt,leftmargin=25pt,rightmargin=25pt}

\title{High-Fidelity Solar Power Income Modeling for Solar-Electric UAVs: Development and Flight Test Based Verification}
\author{
\begin{minipage}{35em}
\begin{center}
Philipp Oettershagen\\
\end{center}
\end{minipage} \\
\\
Autonomous Systems Lab\\Swiss Federal Institute of Technology Zurich (ETH Zurich)\\Leonhardstrasse 21\\ 8092 Zurich\\+41 44 632 7395\\
\texttt{philipp.oettershagen@mavt.ethz.ch} \\
}

\begin{document}

\begin{center}
\Large Technical Report
\end{center}

\maketitle

\vspace{15em}

\begin{abstract}
Solar power models are a crucial element of solar-powered UAV design and performance analysis. During the conceptual design phase, their accuracy directly relates to the accuracy of the predicted performance metrics and thus the final design characteristics of the solar-powered UAV. Likewise, during the operations phase of a solar-powered UAV accurate solar power income models are required to predict and assess the solar power system performance. However, the existing literature on solar-powered UAVs uses highly simplified solar power models. This technical report therefore, first, introduces a high-fidelity solar power model that takes into account the exact aircraft attitude, aircraft geometry, and physical effects such as temperature and the sun radiation's angle-of-incidence that influence the overall solar power system efficiency. Second, models that require a reduced set of input data and are thus more appropriate for the initial design phase of solar-powered UAVs are derived from the high-fidelity model. Third, the models are compared and verified against flight data from a 28-hour continuous day/night solar-powered flight. The results indicate that our high-fidelity model allows a prediction of the average solar power income with an error of less than \unit[5]{\%} whereas previous models were only accurate to about \unit[18]{\%}.
\end{abstract}

\newpage

\section{Introduction}
\label{sec:Introduction}


This technical report extends our previous work on solar power models presented in \parencite{Oettershagen_AeroConf2016}. It is released in conjunction with a publication on the conceptual design, detailed design and flight testing of ETH Zurich's \emph{AtlantikSolar} UAV, a small-scale solar-powered UAV that recently performed an 81-hour continuous flight \parencite{Oettershagen_JFR81hFlight} that is the current \emph{world record} in flight endurance for all aircraft below \unit[50]{kg} total mass. The design- and verification-methods presented in these papers and in similar literature all require a solar power model to predict the amount of electric power $P_\text{solar}$ that the UAV's solar modules collect. In the case of the conceptual design, the incoming solar power is then used to predict the central UAV performance parameters such as the flight endurance $T_\text{endur}$ or, if perpetual flight is feasible, the excess time $T_\text{exc}$. The quality of the solar power model therefore directly relates to the quality of these performance predictions.

\subsection{A basic solar power model}
\label{sec:Introduction_BasicSolarPowerModel}

\textcite{Stein_PVLIB} provides an excellent overview over the physical- and electrical-effects that should be included in precise solar power models. However, the solar power models employed in the current solar-powered UAV design literature are relatively simple. For example, the authors in \parencite{Shiau_OptimalSizingSolarUAV} consider only the average solar radiation over time. A second group of papers such as \parencite{Noth_PhD, Morton_ICRA2013, Klesh_SolarPowerAircraftEnergyOptimalPlanning, Oettershagen_ICRA2015} does model time effects that arise primarily due to the daily solar cycle. A common mathematical model for the instantaneously collected solar power that is also used in our previous work~\parencite{Oettershagen_ICRA2015} is
	\begin{equation} \label{eqn:P_solar}
	P_\text{solar}^\text{nom}=I_\text{solar}(\varphi_\text{lat},h,\delta,t,\vec{n}_\text{sm})\cdot A_\text{sm}\cdot\eta_\text{sm}\cdot\eta_\text{mppt}\;.
	\end{equation}
Here, $I_\text{solar}(\varphi_\text{lat},h,\delta,t,\vec{n}_\text{sm})$ is the solar radiation on a unit (1m\textsuperscript{2}) area that is modeled after \parencite{Duffie_SolarEngineering}. It is a function of geographical latitude $\varphi_\text{lat}$, altitude $h$, current day-of-year $\delta$, local time $t$ and solar module normal vector $\vec{n}_\text{sm}$. For the conceptual design stage of a solar-powered UAV, the solar module area $A_\text{sm}=\text{const}$ is mostly considered a horizontally-oriented area given that the exact orientation depends on the specific mission profile that is only known shortly before the flight operation. Thus, assuming an inertial aircraft-centered North-East-Down (NED) frame of reference, $\vec{n}_\text{sm}=[0,0,-1]$ is chosen. The solar module efficiency $\eta_\text{sm}=\eta_\text{sm}^\text{STC}\cdot\epsilon_\text{sm}^\text{cbr}$ includes an efficiency reduction factor due to the wing camber, and the maximum power point tracker efficiency is $\eta_\text{mppt}$. While the assumptions going into this model are acceptable for the very early solar-powered UAV design stages, they do not allow a sufficiently accurate modeling for the later detailed design and -analysis stages. As an example, consider that a solar-powered UAV shows less-than-expected solar power income during a flight test. The above model is not accurate enough to analyze the issue. Instead, we require a high-fidelity fully time-resolved solar-power model to detect, understand and mitigate these solar power system issues.

\subsection{Contributions of this report}

To answer the need for higher-fidelity solar power income models for solar-powered UAVs, this technical report contributes by

\begin{itemize}
\item Introducing the \emph{Full Solar Power Model} (FM), a fully time-resolved high-fidelity solar power model that takes into account the exact aircraft attitude, aircraft geometry, and physical effects such as temperature and the sun radiation's angle-of-incidence that influence the overall solar power system efficiency. 
\item Deriving the \emph{Conceptual Analysis Model} (CAM) and \emph{Conceptual Design Model} (CDM) from the high-fidelity model. The two models are more appropriate for the initial design phase of solar-powered UAVs because they require less input data (such as certain technological aircraft parameters).  
\item Comparing and verifying all models against flight data from a 28-hour solar-powered flight of the \emph{AtlantikSolar} UAV. The systematic errors are retrieved for each model. The systematic errors can be used by the solar-powered UAV designer to correct the design estimates such that a more precise model-based performance prediction of solar-powered UAVs is possible even in the early design stages.
\end{itemize}

The contributions of this report mean that a solar-powered UAV designer
\begin{itemize}
\item Has the opportunity to select the most appropriate model to work with in a specific development stage (conceptual design, conceptual analysis or detailed flight-test based analysis). 
\item Knows what physical effects (e.g. the angle-of-incidence dependence of the solar module efficiency or the solar module temperature) are driving the solar power system performance. 
\item Is provided information about the prediction accuracy that can be expected from each of these models.
\end{itemize}

\section{A High-Fidelity Solar Power Model} \label{sec:ModelExt_SolarPower}

This section introduces a fully time- and aircraft-state dependent solar power model --- the Full Solar Power Model (FM) in short. It allows to, first, more accurately predict aircraft performance in a specific mission (day-of-year $\delta$, time of day $t$, latitude $\varphi_\text{lat}$, chosen flight path), and second, to detect, understand and mitigate solar power system anomalies. We also use the model to provide a categorization of physical effects that degrade solar module and thus -UAV performance, resulting in \emph{lessons learned} for future UAV designers.

\subsection{Geometric Modelling} \label{sec:ModelExt_SolarPower_Geometry}

\begin{figure}[hbt]
    \centering
    \includegraphics[width=0.8\linewidth]{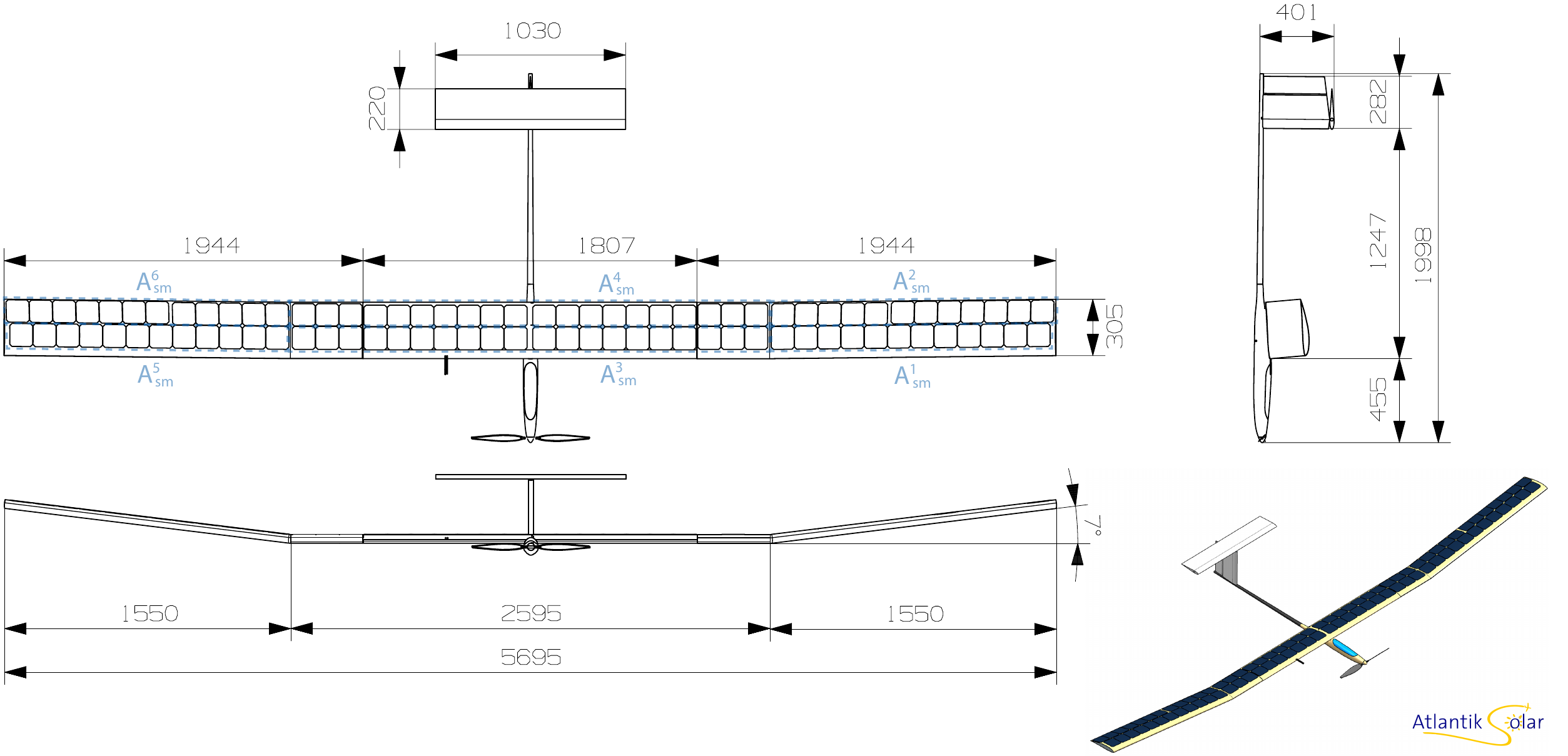}
    \caption{The \emph{AtlantikSolar} UAV airframe. Dimensions are given in \unit{mm}. The solar module geometry of the surfaces $A_\text{sm}^i$ is given in light blue. Image taken from \parencite{Oettershagen_JFR81hFlight}.}
    \label{fig:DetailedDesign_CAD_Airframe}
\end{figure}

\begin{figure}[hbt]
    \centering
     \includegraphics[width=0.5\linewidth]{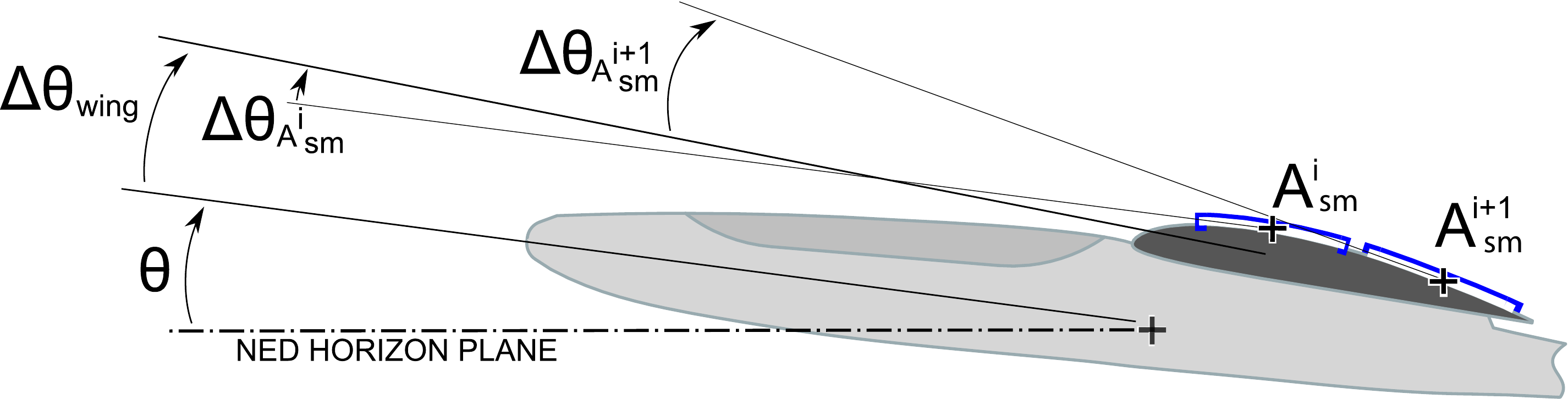}
    \caption{Geometric arrangement of solar module surfaces in the full solar power model}
    \label{fig:SolarPowerModelGeometry}
\end{figure}

We first define an extended geometric representation to calculate the solar power income separately for each of the aircraft's non-horizontal solar modules. Note that we use the \emph{AtlantikSolar} UAV (Figures \ref{fig:DetailedDesign_CAD_Airframe} and \ref{fig:SolarPowerModelGeometry}, with further details given in \parencite{Oettershagen_JFR81hFlight}) as an example for implementing the extended geometric model in this paper. For each of the $N$ solar module-covered areas $A_\text{sm}^i$ we define the orientation relative to the aircraft body frame $\pazocal{B}$ through the surface-normal vector 
	\begin{equation} \label{eqn:SurfaceNormalVec}
	[\vec{n}_\text{sm}^{\;i}]^\pazocal{B} = [\vec{n}_\text{sm}^{\;i}]^\pazocal{B}(\Delta\phi_\text{dih}, \Delta\theta_\text{wing},\Delta\theta_{A_\text{sm}^i}) \, .
	\end{equation}
In lateral direction, the surface normal vector is determined by the wing dihedral angle $\Delta\phi_\text{dih}$. In longitudinal direction, due to the wing upper surface profile every $A_\text{sm}^i$ is mounted at an additional solar cell pitch angle $\Delta\theta_{A_\text{sm}^i}$ with respect to the wing chord line. The relative pitch orientation between the aircraft longitudinal or x-axis (which is approximately aligned with the aircraft's Inertial Measurement Unit longitudinal axis) and the wing chord line is $\Delta\theta_\text{wing}$ and is added on top. Table \ref{tab:SolarModelParameters} summarizes the numerical values chosen for \emph{AtlantikSolar}. Using the aircraft attitude (Euler-angles roll $\phi$, pitch $\theta$ and yaw $\psi$) we can then express the orientation of each $A_\text{sm}^i$ in the inertial frame $\pazocal{I}$ through
	\begin{equation}
	[\vec{n}_\text{sm}^{\;i}]^\pazocal{I}= \mathbf{R}_\pazocal{^B}^\pazocal{I}(\phi,\theta,\psi)\cdot[\vec{n}_\text{sm}^{\;i}]^\pazocal{B} \, ,
	\end{equation}
where $\mathbf{R}_\pazocal{^B}^\pazocal{I}$ is the transformation or direction cosine matrix from the aircraft body to the inertial reference frame. By rewriting Eq.~(\ref{eqn:P_solar}) for all $A^i_\text{sm}$ and performing the corresponding summation, Eq.~(\ref{eqn:P_solar_mult_1}) yields the total incoming solar power $P_\text{solar}$. Equation~(\ref{eqn:P_solar_mult_2}) furthermore allows to separate the global solar irradiation $I_\text{solar}^i$ into its direct component $I_\text{solar}^{i,\text{dir}}$ and diffuse component $I_\text{solar}^{i,\text{diff}}$ with corresponding efficiencies $\eta_\text{sm}^{i,\text{dir}}$ and $\eta_\text{sm}^{i,\text{diff}}$:
	\begin{align}
	P_\text{solar} = \sum_i^N P_\text{solar}^{i} &= \sum_i^N I_\text{solar}^i(\varphi_\text{lat},h,\delta,t,\vec{n}_\text{sm}^{\;i})\cdot \eta_\text{sm}^i\cdot A_\text{sm}^i \cdot\eta_\text{mppt} \label{eqn:P_solar_mult_1}\\
	&=\sum_i^N \Big[I_\text{solar}^{i,\text{dir}}(\varphi_\text{lat},h,\delta,t,\vec{n}_\text{sm}^{\;i})\cdot\eta_\text{sm}^{i,\text{dir}}+I_\text{solar}^{i,\text{diff}}(\varphi_\text{lat},h,\delta,t)\cdot\eta_\text{sm}^{i,\text{diff}}\Big]\cdot A_\text{sm}^i\cdot\eta_\text{mppt}\; \label{eqn:P_solar_mult_2}.
	\end{align}
The extended geometric model also allows to consider shading effects which can degrade module performance for certain aircraft configurations at low sun inclination. It considers shading by the horizontal and vertical tail plane under a sun vector $\vec{r}_\text{sun}$ using a straightforward ray tracing method (see Figure \ref{fig:SolarModel_BlockageSituation}) and yields the shaded area per solar module $A_\text{sm,shaded}^i$. The method does allow a first qualitative assessment, but it does not provide a quantitative measure for the power loss. Calculating the actual power loss based on the model output is possible, but requires modeling the cell-cell and cell-diode interactions within the respective solar module and thus requires knowledge of the specific solar-cell and diode configuration on the airplane. While no generic model can therefore be given here, solar-powered UAV designers are encouraged to implement their own custom plugins to calculate the quantitative power loss.

\begin{table}[hbt]
\caption{Solar model geometric parameters for \emph{AtlantikSolar} AS-2. Square brackets denote sets of variables, in this case for the different solar modules distributed along \emph{AtlantikSolar}'s wing.}
\label{tab:SolarModelParameters}
\begin{center}
\begin{tabular}{l l l}
\toprule
Parameter & Value & Source\\
\midrule
$\Delta\phi_\text{dih}$& $\unit[6.0]{\degree}$ &Aircraft specs\\
$\Delta\theta_\text{wing}$& $\unit[5.7]{\degree}$ &Aircraft specs\\
$\Delta\theta_{A^i_\text{sm}}$&\begin{tabular}{@{}l@{}}$\Delta\theta_{A^{[1,3,5]}}=\unit[-0.5]{\degree}$, $\Delta\theta_{A^{[2,4,6]}}=\unit[9.4]{\degree}$\end{tabular}&Measured\\
$A^{i}_\text{sm}$&\begin{tabular}{@{}l@{}}$A^{[1,3,4,6]}=\unit[0.184]{m^2}$, $A^{[2,5]}=\unit[0.307]{m^2}$\end{tabular}&Aircraft specs\\
\bottomrule
\end{tabular}
\end{center}
\end{table}

\subsection{Solar Module Efficiency Modeling} \label{sec:ModelExt_SolarPower_SolarModuleEfficiency}

\begin{figure}[hbt]
    \centering
    \includegraphics[width=\linewidth]{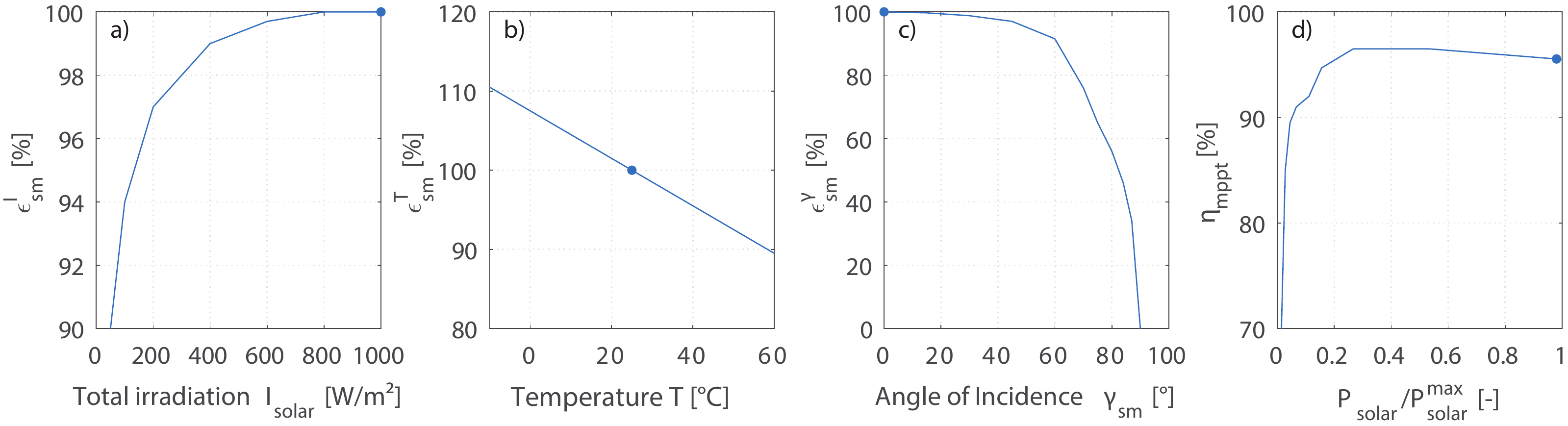}
    \caption{Component efficiencies used in the solar power model. Blue dots represent standard test conditions (STC). }
    \label{fig:SolarModel_EpsilonCurves}
\end{figure}

The standard test conditions (STC) under which solar modules provide their nominal efficiency $\eta_\text{sm}^\text{nom}$ are a total (direct + diffuse) irradiation $I_\text{solar}^\text{STC}=\unitfrac[1000]{W}{m^2}$, a temperature $T_\text{sm}^\text{STC}=\unit[25]{\degree C}$ and an angle-of-incidence $\gamma_\text{sm}^\text{STC}=\unit[0]{\degree}$. Deviations from the STC or nominal conditions will cause efficiency changes that are modeled via the three respective component efficiencies 
	\begin{align}
	\epsilon_\text{sm}^{I} &= \frac{\eta_\text{sm}}{\eta_\text{sm}^\text{STC}(I_\text{solar}=I_\text{solar}^\text{STC})}\bigg|_{T,\gamma = \text{const}} \\
	\epsilon_\text{sm}^{T} &= \frac{\eta_\text{sm}}{\eta_\text{sm}^\text{STC}(T_\text{sm}=T_\text{sm}^\text{STC})}\bigg|_{I,\gamma = \text{const}} \\
	\epsilon_\text{sm}^{\gamma} &= \frac{\eta_\text{sm}}{\eta_\text{sm}^\text{STC}(\gamma_\text{sm}=\gamma_\text{sm}^\text{STC})}\bigg|_{I,T = \text{const}} \;\; .
	\end{align}
The exact correlation for $\epsilon_\text{sm}^{I}(I_\text{solar})$ is retrieved from \parencite{SunPower_YieldReport} and is shown in Figure \ref{fig:SolarModel_EpsilonCurves}a). It mostly results in a relative decrease of $\eta_\text{sm}$ because the total solar irradiation per unit area is usually lower than under standard test conditions. Second, solar module efficiency losses at higher temperatures are represented through the linear relationship
	\begin{equation} \label{eqn:EtaSolarModule}
	\epsilon_\text{sm}^T = 1-c_\text{l}\cdot(T_\text{sm}-T_\text{sm}^\text{STC}) \;,
	\end{equation}
where the loss factor $c_\text{l}=\unitfrac[0.3]{\%}{K}$, applicable to \emph{AtlantikSolar}'s SunPower E60 solar cells, is extracted from solar cell manufacturer data sheets~\parencite{SunPower_E60Datasheet}. The resulting correlation is shown in Figure \ref{fig:SolarModel_EpsilonCurves}b). The instantaneous solar module temperature is approximated using flight test data and the linear relationship
	\begin{equation} \label{eqn:TempSolarModule}
	T_\text{sm}=T_\text{amb}+\Delta T_\text{max} \cdot \frac{P_\text{solar}}{P_\text{solar}^\text{max}}.
	\end{equation}
Here, $T_\text{amb}$ is the ambient temperature measured by the airplane, and $\Delta T_\text{max}\approx \unit[12]{\degree C}$ is the temperature difference between solar module and ambient temperature that was measured in flight at cruise speed approximately at maximum insolation (i.e. for \emph{AtlantikSolar} at $P_\text{solar}^\text{max}\approx \unit[265]{W}$). Third, the exact angle-of-incidence component efficiency $\epsilon_\text{sm}^\gamma(\gamma_\text{sm}^i)$ for the \emph{AtlantikSolar} UAV modules is taken from \parencite{SunPower_YieldReport} and is given in Figure \ref{fig:SolarModel_EpsilonCurves}c). The data shows the expected relative loss of efficiency due to increased reflection that can be of significant importance at high angles of incidence. However, it should be noted that while \parencite{SunPower_YieldReport} consider the same solar cell type, they provide data for a glass front cover and not the specific foil front cover used on the \emph{AtlantikSolar} UAV. The data thus has to be considered a first-order approximation. To determine the angle-of-incidence $\gamma_\text{sm}^i$ for each solar module surface, we use the unit-vectors $\vec{n}_\text{sm}^{\;i}$ and $\vec{r}_\text{sun}$ from the previous section and define
	\begin{equation}
	\gamma_\text{sm}^i = |\arccos(\vec{n}_\text{sm}^{\;i}\cdot \vec{r}_\text{sun})| \; .
	\end{equation}
Together with the wing camber component efficiency $\epsilon_\text{sm}^\text{cbr}$ already introduced in section \ref{sec:Introduction_BasicSolarPowerModel}, we can now express (omitting the surface-index i for clarity) the overall solar module efficiency used in Eq.~(\ref{eqn:P_solar_mult_2}) separately for direct and diffuse radiation as
	\begin{align} 
	\eta_\text{sm}^\text{dir} &= \eta_\text{sm}^\text{STC} \cdot \epsilon_\text{sm}^{I} \cdot \epsilon_\text{sm}^{T}\cdot\epsilon_\text{sm}^{\gamma,\text{dir}}\cdot\epsilon_\text{sm}^\text{cbr} \label{eqn:eta_sm1}\\
	\eta_\text{sm}^\text{diff} &= \eta_\text{sm}^\text{STC} \cdot \epsilon_\text{sm}^{I} \cdot \epsilon_\text{sm}^{T}\cdot \epsilon_\text{sm}^{\gamma,\text{diff}} \;.\label{eqn:eta_sm2}
	\end{align}
Here, $\epsilon_\text{sm}^{\gamma,\text{diff}}=0.83$ is a constant factor that approximates the respective component efficiency over the varying incidence angles in the diffuse radiation. It is retrieved by assuming a uniform distribution of the diffuse radiation $I_\text{diff}$ over the sky's semi-sphere and by performing area-weighed averaging of $\epsilon_\text{sm}^\gamma(\gamma_\text{sm})$ over that area. If no separate calculation of direct- and diffuse-efficiencies is performed (i.e. if Eq.~\ref{eqn:P_solar_mult_1} is used), then we use Eq.~(\ref{eqn:eta_sm1}) and $\eta_\text{sm}=\eta_\text{sm}^\text{dir}$.

\subsection{Maximum Power Point Tracker Efficiency Modeling} \label{sec:ModelExt_SolarPower_MPPTEfficiency}
The maximum power point tracker efficiency $\eta_\text{mppt}$ is a function of the solar power input $P_\text{solar}$ to the MPPTs. The relationship was measured (Figure \ref{fig:SolarModel_EpsilonCurves} d) for the \emph{AtlantikSolar} MPPT in a laboratory test setup. The MPPT performance agrees well with the data sheet values. It reaches efficiencies of $\eta_\text{mppt}^\text{max}=97\%$ but drops to 70\% efficiency at low solar power income. However, during the major part of the power income range, the assumption $\eta_\text{mppt}=95\%$ used for the conceptual design phase is a good fit.

\section{Derivation of Simplified Models for Conceptual UAV Design}
\label{sec:ModelExt_SolarPower_DerivationSimpleModels}

The full solar power model (FM) of the previous section requires the current aircraft and thus solar module attitudes $\vec{n}_\text{sm}^i$, the irradiation level $I_\text{solar}$ and the current solar module temperature $T_\text{sm}$ as input parameters. However, quantities $\vec{n}_\text{sm}^i$ and $T_\text{sm}$ are not available during the UAV conceptual design and analysis phase because the exact operating conditions (e.g. the chosen flight path) are unknown. Therefore, Table \ref{tab:SolarPowerModels} introduces models that run with a less extensive set of input data. These are the
\begin{itemize}
	\item \emph{Conceptual Design Model} (CDM), which is and was used in the conceptual design of \emph{AtlantikSolar} in \parencite{Oettershagen_JFR81hFlight}. It assumes a single, flat and always horizontal solar module surface, a constant MPPT efficiency $\eta_\text{mppt}$ and a constant solar module efficiency $\eta_\text{sm}$ that is only a function of a user specified constant temperature $T_\pazocal{C}$. 
	\item \emph{Conceptual Analysis Model} (CAM), which adds the angle-of-incidence correlation of Figure \ref{fig:SolarModel_EpsilonCurves}c) to obtain a time-varying efficiency $\eta_\text{sm}$. It is used for a more detailed performance prediction once technical parameters such as the $\gamma_\text{sm}$-correlation are known, but the exact conditions such as the flight path, wind conditions and thus flight attitude are still unknown. This is often the case during the late conceptual design phase and the detailed design phase. In \parencite{Oettershagen_JFR81hFlight} the CAM is used to provide an accurate performance outlook after the design and flight verification has been completed.
	\item \emph{Verification Model} (VM), which compensates for offsets in the required heading and roll angles (especially in a loitering mission) caused by horizontal winds. The VM approximates the aircraft geometry with a single flat surface with the time-varying and wind-dependent orientation of the central wing $\vec{n}_\text{sm}^\text{CtrWing}$. The VM is not used for actual solar-powered UAV design or analysis, but only used internally in this report to assess the magnitude of the CDM/CAM modeling errors that are \emph{not} caused by wind using `wind-disturbed flight test data'. 
\end{itemize}

\begin{table}[htb] 
\caption{Overview over the solar power models proposed for different stages of UAV development. The subscript $\pazocal{C}$ denotes a constant value. The last row indicates whether $P_\text{solar}$ is calculated using separate efficiencies for the direct and diffuse radiation.} 
\label{tab:SolarPowerModels}
\begin{center}
\begin{tabular}{l l l l l}
\toprule
  Name & CDM & CAM & VM & FM\\ 
  Application & Conc. Design & Conc. Analysis & Verification (of CA) & Full model\\ 
\midrule
 Surfaces & Single & Single & Single & Multiple\\
 Geometry & Flat & Flat & Flat & Realistic\\
 Attitude & const (horiz.) & const (horiz.) & $f(\vec{n}_\text{sm}^\text{CtrWing})$ & $f(\vec{n}_\text{sm}^i)$\\
$\eta_\text{mppt}$ & const & const & const & $f(P_\text{solar})$ \\
$\eta_\text{sm}$ & $f(T_\pazocal{C})$ & $f(\gamma_\text{sm}(\vec{n}_\pazocal{C}),T_\pazocal{C})$ & $f(\gamma_\text{sm}(\vec{n}_\text{sm}^\text{CtrWing}),T_\pazocal{C})$ & $f(\gamma_\text{sm}^i(\vec{n}_\text{sm}^i),I_\text{solar},T_\text{sm})$ \\
 $P_\text{solar}$ & eqn. (\ref{eqn:P_solar_mult_1}) & eqn. (\ref{eqn:P_solar_mult_2}) & eqn. (\ref{eqn:P_solar_mult_2}) & eqn. (\ref{eqn:P_solar_mult_2})\\ 
\bottomrule
\end{tabular}
\end{center}
\end{table}

\section{Solar Power Model Verification}
\label{sec:ModelExt_SolarPower_VerificationAllModels}

This report assesses the developed solar power models using flight data from the 28-hour solar-powered flight performed by \emph{AtlantikSolar} AS-2 that is described in more detail in \parencite{Oettershagen_AeroConf2016}. Note that this flight only includes a single day/night solar cycle. A verification that uses the more comprehensive dataset from the three day/night cycles of \emph{AtlantikSolar}'s 81-hour flight is performed (though in less detail) in \parencite{Oettershagen_JFR81hFlight}.

\subsection{Full Solar Power Model Verification} 
\label{sec:ModelExt_SolarPower_FMVerification}
    
Figure \ref{fig:SolarModel_DetailedPlots} compares modeled and measured solar power system data for three segments (early- to late-morning) of the aforementioned 28-hour flight. To assess the accuracy of the developed solar power income model, we employ the standard definitions for the absolute average-, RMS- and maximum model error $\hat{E}_x$ (where a negative error represents an underestimate by the model). In addition, we define the respective relative error of quantity $x$ as $\hat{e}_x=\nicefrac{\hat{E}_x}{x_\text{avrg}^\text{exp}}$. The three time segments of the aforementioned flight exhibit characteristically different solar power system behavior. More specifically:

\begin{figure}[p]
    \centering
    \includegraphics[width=\linewidth]{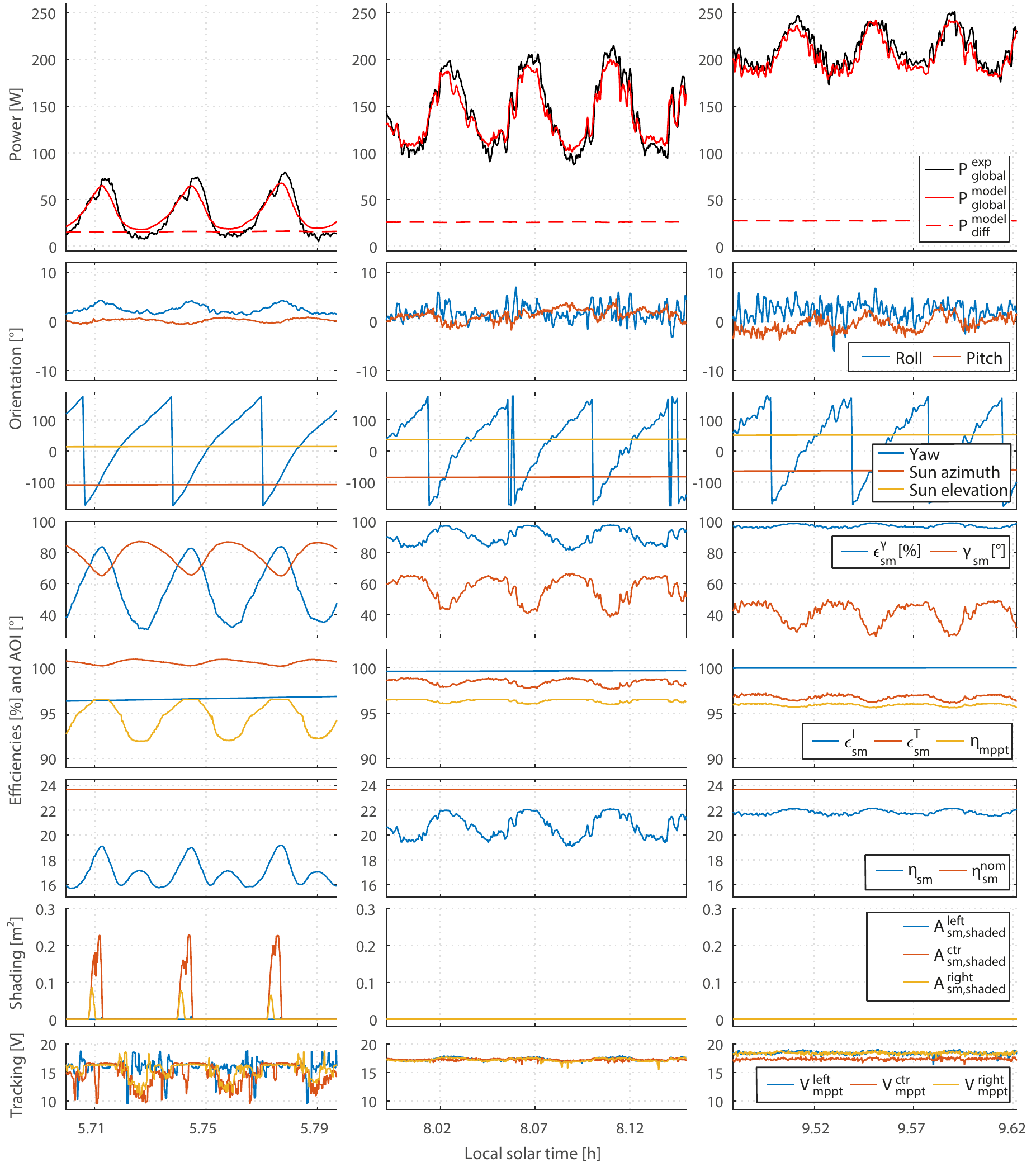}
    \caption{Comparison of results from our full solar power model and test flight data collected during the second day of a continuous 28-hour flight of \emph{AtlantikSolar} AS-2 on July 1\textsuperscript{st} 2015. The three patches show early-morning (left), mid-morning (center) and late-morning (right) flight. The graphs show experimental ($P_\text{solar}^\text{exp}$) and modeled ($P_\text{solar}^\text{model}$) solar power income, measured aircraft orientation, modeled component and total efficiencies, modeled shaded solar module area and the measured maximum power point tracker voltage as an indicator of solar power system tracking behavior. The data is recorded at $\unit[2]{Hz}$ and a two-sided moving average filter with a semi window length of 2 samples is applied. The model efficiency and angle data is plotted as averages over the six solar module sections of the UAV.}
    \label{fig:SolarModel_DetailedPlots}
\end{figure}

\begin{figure}[htb]
    \centering
    \includegraphics[width=0.8\linewidth]{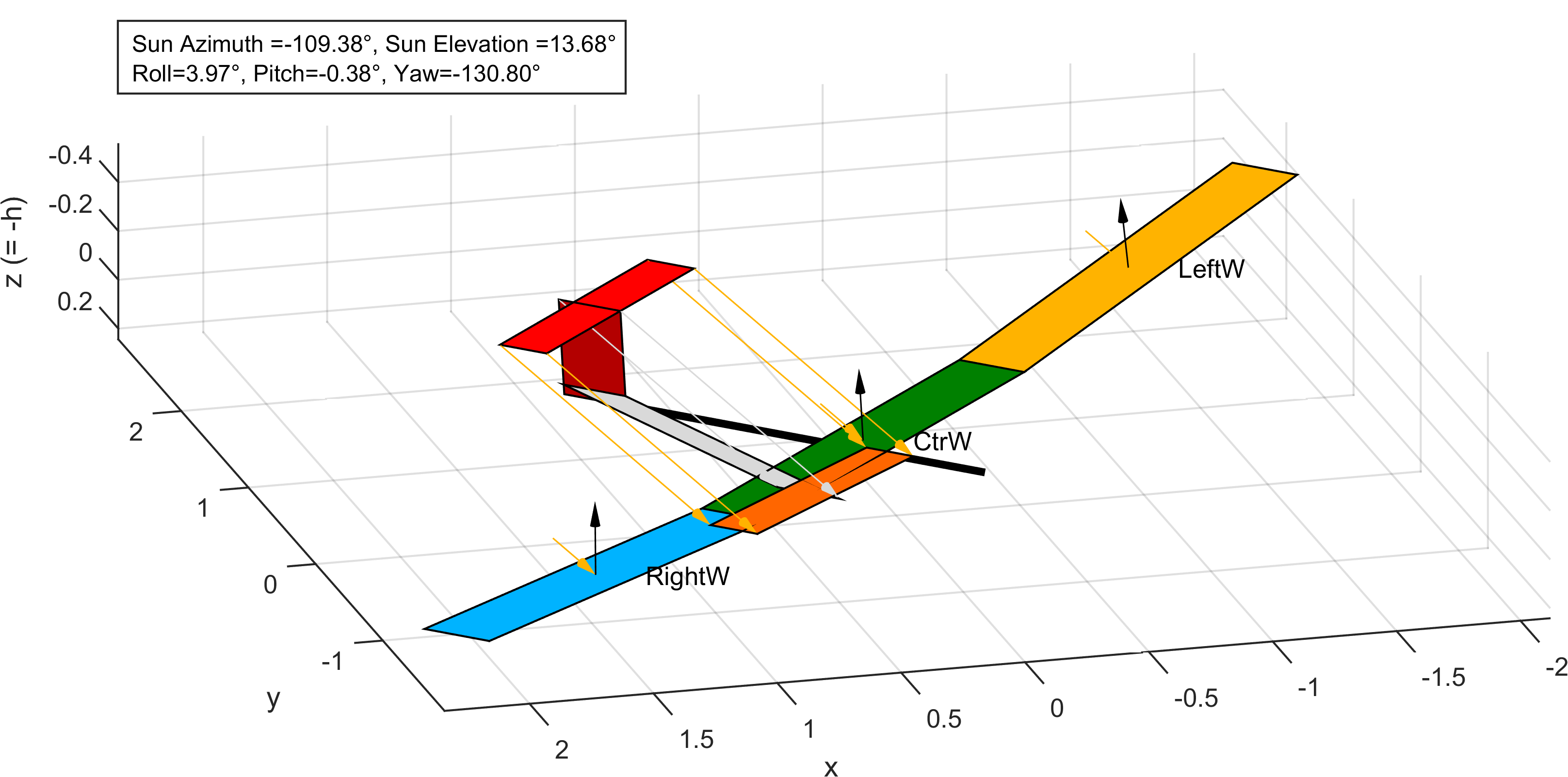}
    \caption{Visualization of the full model geometry and incoming sun irradiation at solar time $t=\unit[5.71]{h}$ of \emph{AtlantikSolar}'s 28-hour flight. The black arrows are the surface normal vectors $\vec{n}_\text{sm}^i$ of each wing, the orange arrows are sun rays and can be used to infer the angles of incidence $\gamma_\text{sm}^i$. The grey and dark-orange surfaces are the shadows of the rudder and elevator projected onto the solar-module covered wing.}
    \label{fig:SolarModel_BlockageSituation}
\end{figure}

\begin{itemize}
\item During \emph{early-morning flight} (Figure \ref{fig:SolarModel_DetailedPlots}, left) at $t=[\unit[5.71]{h}, \unit[5.79]{h}]$ local solar time, the low sun elevation causes a significant variation of the solar irradiation levels with the aircraft yaw angle. Both $\epsilon_\text{sm}^\gamma$ and $\eta_\text{mppt}$ therefore play a significant role, while $\epsilon_\text{sm}^I$ and $\epsilon_\text{sm}^T$ only have a minor effect. The resulting overall module efficiency $\eta_\text{sm}$ is only 16-18\% compared to the standard operating condition efficiency of $\eta_\text{sm}^\text{STC}=\unit[23.7]{\%}$. The model errors are $\hat{E}^\text{RMS}_{P_\text{solar}}=\unit[8.54]{W}$, $\hat{e}^\text{RMS}_{P_\text{solar}}=\unit[25.4]{\%}$, $\hat{e}^\text{max}_{P_\text{solar}}=\unit[43.4]{\%}$ and $\hat{e}^\text{avrg}_{P_\text{solar}}=\unit[5.14]{\%}$ ($P_\text{solar}^\text{model}=\unit[35.3]{W}$ and $P_\text{solar}^\text{exp}=\unit[33.6]{W}$). Two graphs help to explain this overestimation by the model: First, the recorded MPPT voltages (Figure \ref{fig:SolarModel_DetailedPlots}, bottom) indicate that in low irradiation conditions (i.e. the solar panels face away from the sun), the employed MPPTs fail to track the solar panel maximum power point. This clearly indicates sub-optimal power system behavior that, especially for flights in low total radiation or low sun elevation conditions, needs improvement. Second, the power plots show small notches shortly before the model predicts maximum solar power income. Our geometric airplane model indicates (Figure \ref{fig:SolarModel_DetailedPlots}, bottom) that shading of the center and right solar modules by the aircraft tail exactly at $t=\unit[5.71]{h}$ can explain this. Figure \ref{fig:SolarModel_BlockageSituation} represents the output of the shading calculations in the full solar power model at this time.

\item During \emph{mid-morning flight} (Figure \ref{fig:SolarModel_DetailedPlots}, center) at $t=[\unit[8.02]{h},\unit[8.12]{h}]$, the average solar power income has increased considerably to $P_\text{solar}^\text{model}=\unit[140.5]{W}$ and $P_\text{solar}^\text{exp}=\unit[142.1]{W}$. The average error is only $\hat{e}^\text{avrg}_{P_\text{solar}}=\unit[-1.1]{\%}$, and the RMS and maximum errors have decreased to $\hat{E}^\text{RMS}_{P_\text{solar}}=\unit[11.3]{W}$, $\hat{e}^\text{RMS}_{P_\text{solar}}=\unit[7.98]{\%}$ and $\hat{e}^\text{max}_{P_\text{solar}}=\unit[21.3]{\%}$. Due to the higher sun elevation, the variations in $\epsilon_\text{sm}^\gamma$, $\epsilon_\text{sm}^\gamma$ and $\eta_\text{mppt}$ are much less pronounced. Solar module shading and sub-optimal MPP-tracking are not observed anymore.
 
\item During \emph{late-morning flight} (Figure \ref{fig:SolarModel_DetailedPlots}, right) at $t=[\unit[9.52]{h},\unit[9.62]{h}]$ local time we retrieve $P_\text{solar}^\text{model}=\unit[204.7]{W}$ and $P_\text{solar}^\text{exp}=\unit[211.5]{W}$. With an estimated $\epsilon_\text{sm}^T\approx\unit[96]{\%}$, the elevated temperature of $T_\text{sm}=\unit[37]{\degree C}$ at $T_\text{amb}=\unit[27]{\degree C}$ now noticeably reduces the solar module efficiency. Despite the approaching solar irradiation maximum, $\gamma_\text{sm}$ remains between 31\degree and 49\degree, thus underlining the importance of considering an exact $\epsilon_\text{sm}^\gamma$ correlation (Figure \ref{fig:SolarModel_EpsilonCurves}) even at high sun elevation. The RMS and maximum errors are $\hat{E}^\text{RMS}_{P_\text{solar}}=\unit[9.39]{W}$, $\hat{e}^\text{RMS}_{P_\text{solar}}=\unit[4.4]{\%}$ and $\hat{e}^\text{max}_{P_\text{solar}}=\unit[9.9]{\%}$ respectively. The average error has increased in magnitude to $\hat{e}^\text{avrg}_{P_\text{solar}}=\unit[-3.2]{\%}$. The full solar power model thus tends to underestimate the incoming solar radiation for high irradiation conditions. 
\end{itemize}

\begin{figure}[!b]
    \centering
    \includegraphics[width=\linewidth]{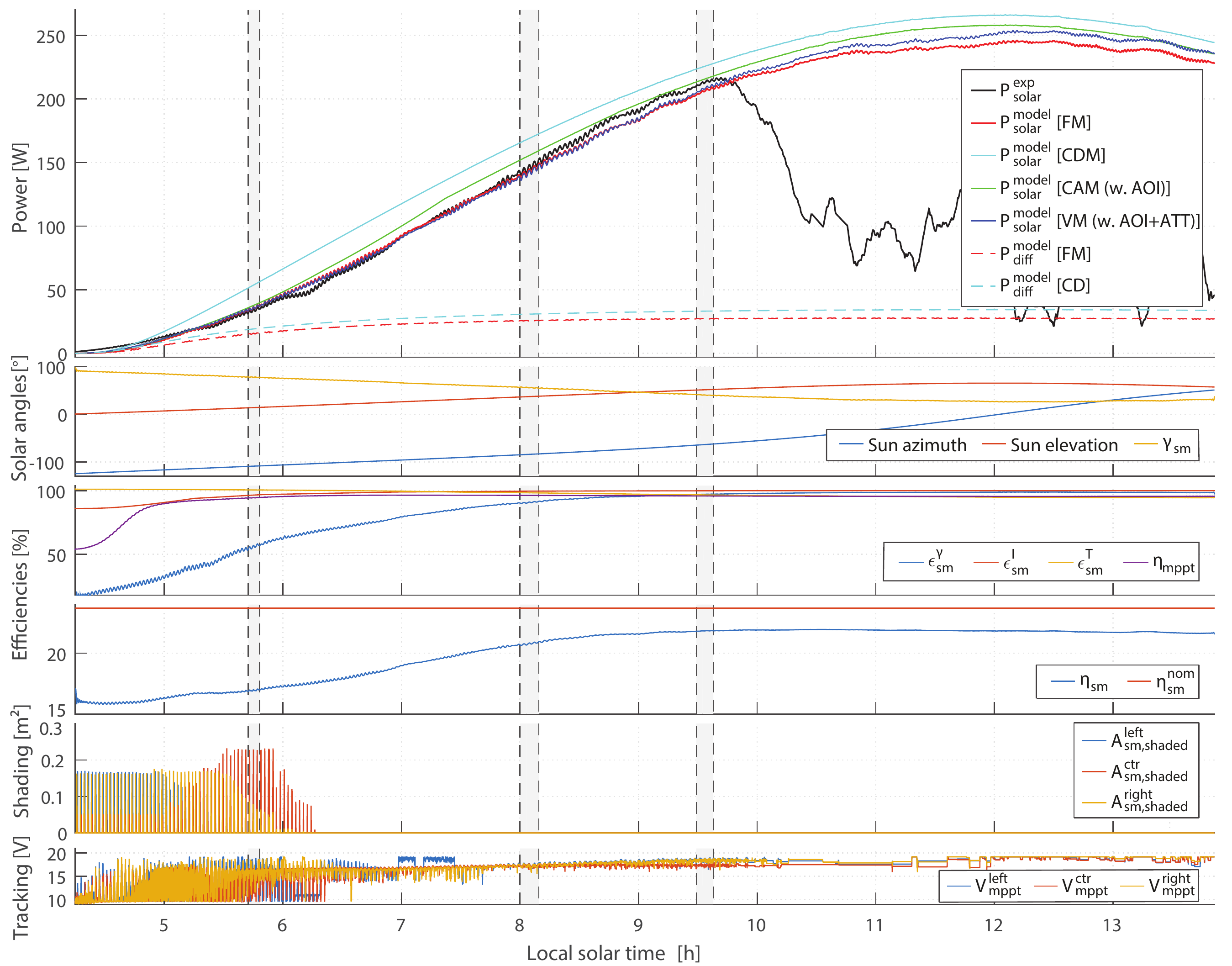}
    \caption{Comparison of the solar power models of Table \ref{tab:SolarPowerModels} and flight data. The data is recorded at $\unit[2]{Hz}$ and, except for the shading and tracking plots, a two-sided moving average filter with a semi window length of 800 samples is applied. The grey patches represent the detailed plots of Figure \ref{fig:SolarModel_DetailedPlots}. The model efficiency and angles are averages over all UAV solar modules. The decrease in $P_\text{solar}^\text{exp}$ at $t=\unit[9.72]{h}$ is only because the batteries are nearly full and the MPPTs thus reduce their power output.}
    \label{fig:SolarModel_OverviewPlots}
\end{figure}

A comprehensive assessment over a full day (Figure \ref{fig:SolarModel_OverviewPlots}) confirms the overestimation of $P_\text{solar}$ during the early morning and the underestimation during the higher radiation conditions of the late morning. The plots clarify that solar module shading occurs from sunrise to around $t=\unit[6.26]{h}$ (or more generally until the sun elevation angle is above $\unit[19]{\degree}$ for this UAV geometry) and accounts for less than a 10\% decrease in $P_\text{solar}$ even during the most significant shading problems around $t=\unit[5.71]{h}$. Although shading by \emph{AtlantikSolar's} T-tail is not significant over the full day in the presented conditions, aircraft designed for optimum performance at low sun elevation conditions could consider alternative tail configurations (e.g. an inverted T-tail) or optimized path planning to reduce the effects of shading. In contrast to shading, the imperfect maximum power point tracking under low light conditions is found to occur until $\unit[7.77]{h}$ and thus plays an even more significant role. The discrepancies in the high-radiation conditions around noon are expected to lie in the first-order approximation for the $\epsilon_\text{sm}^\gamma$- and $\epsilon_\text{sm}^T$-correlations. Removing these deviations would require accurate measurements of $\epsilon_\text{sm}^\gamma$ with high-precision laboratory equipment such as solar flashers that was not available for this paper. Note that to exclude errors in the underlying solar irradiation $I_\text{solar}$, the model by \parencite{Duffie_SolarEngineering} was compared to the \emph{Sandia National Labs} PV\_LIB \emph{toolbox} model \parencite{Stein_PVLIB} and found to agree well within 3\% maximum error. 

Overall, given all these uncertainties, the full model accuracy is satisfactory: The average deviation over the domain in which the MPPTs operate without limiting their output current ($t=[\unit[4.17]{h},\unit[9.72]{h}]$) is $\hat{e}^\text{avrg}_{P_\text{solar}}=\unit[-1.75]{\%}$ ($P_\text{solar}^\text{model}=\unit[97.9]{W}$ and $P_\text{solar}^\text{exp}=\unit[99.6]{W}$). The RMS error is, mostly due to the higher amplitude of the heading-dependent variations of $P_\text{solar}$, $\hat{E}^\text{RMS}_{P_\text{solar}}=\unit[9.84]{W}$ or $\hat{e}^\text{RMS}_{P_\text{solar}}=\unit[9.88]{\%}$. These model errors are relatively small compared to all other possible design errors (caused e.g. by wrong user inputs for subsystem masses as described in \parencite{Oettershagen_JFR81hFlight}), and the full model can thus be considered accurate enough for solar-powered UAV design and performance analysis.

\subsection{Comparison of Full- and Simplified-Models}

Figure \ref{fig:SolarModel_OverviewPlots} and Table \ref{tab:SolarPowerModels_comparison} provide a detailed performance comparison of all models. As stated before, even the full model (FM) shows deviations from the flight test data. The findings for the other models are:

\begin{itemize}
	\item \emph{Verification Model}: Over the full day, the difference between verification- and full model is small. This means that the additional physical effects modeled in the FM do not have a significant influence for the chosen aircraft and power system. Specifically, $\eta_\text{mppt}$ and $\eta_\text{sm}^\text{I}$ mainly deviate from their nominal values in the morning, and the temperature dependence leads to a decreased $P_\text{solar}$ during noon. Consequently --- and this is an important finding --- the conceptual design of solar-powered UAVs, e.g. through the approach presented in \parencite{Oettershagen_JFR81hFlight}, can be safely performed without these additional effects of the FM. 
	\item \emph{Conceptual Analysis Model}: The relative error $\hat{e}_{P_\text{solar}}^\text{avrg}$ differs by around 7 percent points between VM and CAM. The only difference between these models is the inclusion of the current aircraft attitude through $\vec{n}_\text{sm}^\text{CtrWing}$. The deviation is caused by two effects: First, due the concave-downwards nature of $\epsilon_\text{sm}^\gamma(\gamma_\text{sm})$ in Figure \ref{fig:SolarModel_EpsilonCurves}c), oscillations of the \emph{instantaneous} angle-of-incidence $\gamma_\text{sm}(t)$ around an \emph{average} $\gamma_\text{sm}^\text{avrg}=\text{const}$ result in a lower average $P_\text{solar}$ than for $\gamma_\text{sm}(t)=\text{const}=\gamma_\text{sm}^\text{avrg}$. Second, during the loitering turns, a horizontal eastern wind of only $\unitfrac[2]{m}{s}$ causes the solar panels to face away from the sun longer than towards the sun. Consequently, the CAM overestimates $P_\text{solar}$ during the morning. This confirms that wind has a significant influence, and although wind data is usually not available in the conceptual design stage, its effects especially during the morning and evening need to be considered as early as possible. 
	\item \emph{Conceptual Design Model}: The large deviation between CAM and CDM is explained by the inclusion of the  $\gamma_\text{sm}$-dependence of $\eta_\text{sm}$ in the CAM. The respective component efficiency $\epsilon_\text{sm}^\gamma$ is only $\unit[20]{\%}$ around sunrise (see section \ref{sec:ModelExt_SolarPower_FMVerification}). Therefore, $\epsilon_\text{sm}^\gamma$ needs to be estimated and considered as early as possible during the conceptual analysis stage.
\end{itemize} 

\begin{table}[htb]
\caption{Estimation errors of the four solar power models. The subscript $\pazocal{F}$ in $P_{\text{solar},\pazocal{F}}$ denotes that the errors are calculated based on pre-filtered data (using the same filter as Figure \ref{fig:SolarModel_OverviewPlots}), in order to allow a meaningful comparison between the attitude-dependent VM and FM models and the attitude-independent CD and CM models. The application of the filter only influences the RMS errors.}
\begin{center}
\label{tab:SolarPowerModels_comparison}
\begin{tabular}{l l l l l l}
\toprule
  Name & CDM & CAM & VM & FM & Flight Data\\ 
  Application & Conc. Design & Conc. Analysis & Verification (of CA) & Full model &\\ 
\midrule
\addlinespace[0.4em] 
 $P_\text{solar}^\text{avrg}$ & \unit[117.35]{W} & \unit[104.89]{W} & \unit[97.65]{W} & \unit[97.87]{W} & \unit[99.61]{W}\\
\addlinespace[0.4em] 
 $\hat{e}_{P_\text{solar}}^\text{avrg}$ & \unit[17.81]{\%} & \unit[5.31]{\%} & \unit[-1.96]{\%} & \unit[-1.75]{\%} & --- \\
\addlinespace[0.4em] 
 $\hat{E}_{P_{\text{solar},\pazocal{F}}}^\text{RMS}$ & \unit[19.95]{W} & \unit[6.99]{W} & \unit[3.94]{W} & \unit[4.17]{W} & ---\\
\addlinespace[0.4em] 
 $\hat{e}_{P_{\text{solar},\pazocal{F}}}^\text{RMS}$ & \unit[20.03]{\%} & \unit[7.02]{\%} & \unit[3.96]{\%} & \unit[4.19]{\%} & ---\\
\addlinespace[0.2em] 
\bottomrule
\end{tabular}
\end{center}
\end{table}

Overall, the comparison shows that while the full model allows accurate solar power income prediction, the less sophisticated conceptual design (CDM) and analysis models (CAM) show errors that are of significance during the conceptual design. The two most important factors that a designer of solar-powered UAVs should consider as early as possible to decrease these deviations are the angle-of-incidence correlation $\epsilon_\text{sm}^\gamma$ for $\eta_\text{sm}$ and the related effects of changing aircraft attitude and horizontal winds. Both effects increase in importance when operating in low sun elevation angles. 

\section{Conclusion and Lessons Learned}
\label{sec:Conclusion}

Of the models derived in this paper and verified via the solar power data recorded during a 28-hour solar-powered flight, the Full Solar Power Model (FM) provided accurate solar power prediction results with an error in the average $P_\text{solar}$ of \unit[-1.75]{\%}. The model however requires input data that is not available during the conceptual design stage of solar-powered UAVs. The models developed for these purposes, the Conceptual Design Model (CDM) and Conceptual Analysis Model (CAM), have to make simplifications that result in higher errors in the average $P_\text{solar}$. More precisely, the CDM and CAM overestimate $P_\text{solar}$ by approximately \unit[18]{\%} and \unit[5]{\%} respectively. The lessons learned from this analysis that can be used by other solar-powered UAV designers are

\begin{itemize}
\item The Conceptual Design Model (CDM) should only be used in the very early design stage of a solar-powered UAV. The solar power income and thus overall flight endurance of the designed UAV will likely be lower than predicted.
\item The Conceptual Analysis Model (CAM) should replace the CDM as soon as possible, even in the early conceptual design phase of a UAV. The CAM incorporates the angle-of-incidence sensitivity of the solar module efficiency $\eta_\text{sm}$ and thereby reduces the prediction error significantly, i.e. in the specific conditions of the presented 28-hour flight from \unit[18]{\%} to \unit[5]{\%}.
\item The Full Solar Power Model (FM) shall then be used for actual solar power income prediction once the flight operations phase has begun and the flight path and current wind conditions are known. The FM is accurate enough for a detailed solar power system performance analysis, i.e. it can for example be used to verify that a solar power system is working properly.
\end{itemize}

\printbibliography


\end{document}